\documentclass{article}

\usepackage[preprint]{neurips_2023}

\usepackage[utf8]{inputenc} 
\usepackage[T1]{fontenc}    
\usepackage{hyperref}       
\usepackage{url}            
\usepackage{booktabs}       
\usepackage{amsfonts}       
\usepackage{nicefrac}       
\usepackage{microtype}      
\usepackage{xcolor}         

\usepackage{graphicx}
\usepackage{amsmath}
\usepackage[compact]{titlesec}
\titlespacing{\section}{0pt}{0pt}{0pt}
\titlespacing{\subsection}{0pt}{0pt}{0pt}

\newcommand{\cgh}[1]{\textcolor{black}{#1}}
\usepackage{tabularray}
\usepackage{wrapfig}

\title{Towards \cgh{a} Robust Retrieval-\cgh{B}ased Summarization System}

\author{%
Shengjie Liu\thanks{These authors contributed equally to this work.} 
 \\
  North Carolina State University\\
  \texttt{sliu56@ncsu.edu} \\
\And
Jing Wu\protect\footnotemark[1] \\
University of Illinois Urbana-Champaign\\
\texttt{jingwu6@illinois.edu} \\
\And
Jingyuan Bao \\
Northwestern University \\
\texttt{jingyuanbao2018@u.northwestern.edu} \\
\And
Wenyi Wang \\
North Carolina State University\\
\texttt{wwang52@ncsu.edu} \\
\And
Naira Hovakimyan \\
University of Illinois Urbana-Champaign\\
\texttt{nhovakim@illinois.edu} \\
\And 
Christopher G Healey\thanks{The Corresponding Author}\\
North Carolina State University\\
\texttt{healey@ncsu.edu} \\
}

\begin{document}

\maketitle

\begin{abstract}
This \cgh{paper describes an investigation of the robustness of large language models (LLMs) for} retrieval augmented generation (RAG)-based summarization tasks. While LLMs \cgh{provide} summarization \cgh{capabilities}, their performance in complex, real-world scenarios remains \cgh{under-explored}. \cgh{Our first contribution is \textit{LogicSumm}}, an innovative evaluation framework \cgh{incorporating} realistic scenarios to assess LLM robustness \cgh{during} RAG-based summarization. \cgh{Based on limitations identified by \textit{LogiSumm}, we then} developed \textit{SummRAG}, a comprehensive system \cgh{to create} training dialogues \cgh{and} fine-\cgh{tune} a model \cgh{to enhance} robustness within \textit{LogicSumm}'s scenarios. \cgh{\textit{SummRAG} is an example of our goal of defining structured methods to test the capabilities of an LLM, rather than addressing issues in a one-off fashion.} Experimental results confirm the \cgh{power} of \textit{SummRAG}, showcasing \cgh{improved} logical coherence and summarization quality.
Data, corresponding model weights, and Python code are available online$^{1}$.
\end{abstract}

\section{Introduction}

\stepcounter{footnote}
\footnotetext{
\url{https://huggingface.co/datasets/zycjlsj123/ragsummdata}; \url{https://huggingface.co/zycjlsj123/rag_summ}; \url{https://github.com/ncsulsj/Robust_Sumsystem}}
In the evolving landscape of automated text summarization, \cgh{large language models} (LLMs) have emerged as key players, demonstrating remarkable efficiency in distilling complex information into concise summaries. Pioneering works such as those by \cite{goyal2022news,liu2022brio,liu2022revisiting} highlight the \cgh{progress} made in leveraging LLMs for this purpose. \cgh{R}elated benchmarks underscore \cgh{LLMs'} growing significance in this field \cite{zhang2024benchmarking}. Despite these \cgh{advances}, LLMs encounter a critical bottleneck: their training datasets are static, making \cgh{the integration of} new information post-training a formidable challenge. 

Retrieval Augmented Generation (RAG) \cgh{was} introduced to recognize this limitation. By integrating external knowledge sources, LLMs are empowered to \cgh{dynamically} incorporate up-to-date information \cgh{in} real-time \cgh{during} generation tasks \cite{lewis2020retrieval,izacard2022few,guu2020retrieval}. \cgh{RAGs} \cgh{promise} to address the issue of \cgh{a} static knowledge base in \cgh{an} LLM, paving the way for more accurate and up-to-date summaries. 

Although RAG integration with LLMs offers a promising avenue for more comprehensive and current summaries, \cgh{research} specifically focused on summarization using RAG and LLMs is under-explored. This gap manifests in two significant \cgh{limitations}: (1) \textbf{Evaluation Pipeline.} \cgh{T}he absence of targeted evaluation pipelines for assessing this specific use case, and (2) \textbf{Effective Methods.} The scarcity of research directly discussing the application of RAG in conjunction with LLMs for summarization. 

To address \cgh{these} gaps in summarization research using LLMs with RAG, we propose a novel evaluation pipeline \textbf{LogicSumm}. This pipeline is designed to systematically understand and benchmark the summarization capabilities of LLMs augmented with RAG. Our approach includes \cgh{addressing} the most commonly encountered scenarios \cgh{during} summarization, \cgh{split} into seven distinct cases \cgh{and} providing a comprehensive framework for evaluation. We conduct experiments using popular LLMs integrated with RAG across these cases. 

\cgh{Across} our seven cases, we observed a significant performance decline in previous RAG-based summarization approaches \cgh{for} input \cgh{that included} documents \cgh{that were} irrelevant to the topic being summarized. This finding highlights a significant challenge: the difficulty in effectively identifying relevant documents for stable summarization. We \cgh{develop} a novel support system \textbf{SummRAG} \cgh{that constructs data contextually to fine-tune a model and} improve its robustness in all scenarios with \cgh{minimal reliance on external datasets.} This framework boosts the performance of public language models in summarization tasks, effectively narrowing the performance gap with more advanced but less accessible models like GPT-4. \cgh{It also demonstrates one of our motivating objectives: developing structured, generalizable frameworks to address related classes of issues, rather than solving problems in a one-off fashion.}

In summary, our paper \cgh{provides} the following \cgh{novel} contributions.

\begin{enumerate}
    \item  We \cgh{investigate} the important but under-explored domain \cgh{of} RAG-based summarization with LLMs. To the best of our knowledge, we propose the first evaluation pipeline \textit{LogicSumm} \cgh{tested using} seven \cgh{summarization scenarios that} thoroughly assess the summarization capabilities of LLMs under a range of \cgh{common use cases}. 
    
    \item We present \textit{SummRAG}, a comprehensive end-to-end framework that encompasses both dialogue generation and \cgh{model} fine-tuning \cgh{to improve} the robustness and overall performance of RAG-based summarization.
    
    \item We publish a new dataset from SummRAG \cgh{that} is model-agnostic and capable of enhancing public LLMs in scenarios pertinent to RAG-based summarization tasks. 
\end{enumerate}

\section{Related Work}

\subsection{Large Language Model}

The evolution of LLMs began with the advent of transformers \cite{vaswani2017attention}. This development significantly enhanced language models' versatility across various tasks, a breakthrough prominently showcased by BERT \cite{devlin2018bert}. Following these \cgh{advances}, the focus shifted towards the development of larger-scale models informed by the scaling law \cite{kaplan2020scaling}. This led to the creation of groundbreaking models like GPT \cite{brown2020language}, LLaMA \cite{touvron2023llama}, PaLM \cite{chowdhery2023palm}, Jurassic \cite{lieber2021jurassic}, Mistral \cite{jiang2023mistral}, and Claude, characterized by their tens of billions of parameters. These models unlocked advanced in-context learning and zero-shot performance across various tasks.

\subsection{Retrieval Augmented Generation}

Retrieval-augmented generation \cgh{was} introduced as a pivotal enhancement for language models, providing access to a wealth of additional knowledge by retrieving information from external databases \cite{lewis2020retrieval,guu2020retrieval,borgeaud2022improving}. When combined with LLMs, RAG significantly enhances the \cgh{ability} for up-to-date and accurate generation tasks such as open-domain QA \cite{izacard2020leveraging,karpukhin2020dense,guu2020retrieval}, dialogue \cite{cai2018skeleton}, and code generation \cite{parvez2021retrieval}. Certain challenges have also been noted, however. Studies highlight that noise in the retrieved text can adversely affect the performance of the language model, potentially leading to misinformation or errors \cite{chen2023benchmarking,xu2024list,xu2023search}. There is also \cgh{the potential for} conflict between user-provided \cgh{text} and \cgh{information} retrieved by RAG \cite{jin2024tug}. These issues underscore the necessity \cgh{to develop} a more refined framework that enhances both the robustness and consistency of LLM-based RAG systems.

\subsection{Text Summarization}

Text summarization involves condensing the core content of a \cgh{text document} into a concise \cgh{summary}, extracting and synthesizing key points to accurately represent the original article \cite{nenkova2011automatic, chen2018fast}. Traditional approaches have utilized methods based on word frequency to determine salience \cite{nenkova2006compositional} and explored discourse semantics \cite{steinberger2007two}. Fine-tuning techniques have also been applied \cite{liu2019text,lewis2019bart, zhang2020pegasus, liu2022brio}. More recently, LLMs have emerged as a central component in text summarization, significantly \cgh{impacting} the development and effectiveness of summarization techniques \cite{zhang2024benchmarking, tang2023evaluating,van2023clinical}.

\section{LogicSumm}
\label{sec: logicsumm}

\begin{figure}[!t] 
\centering 
 \includegraphics[width=1\linewidth]{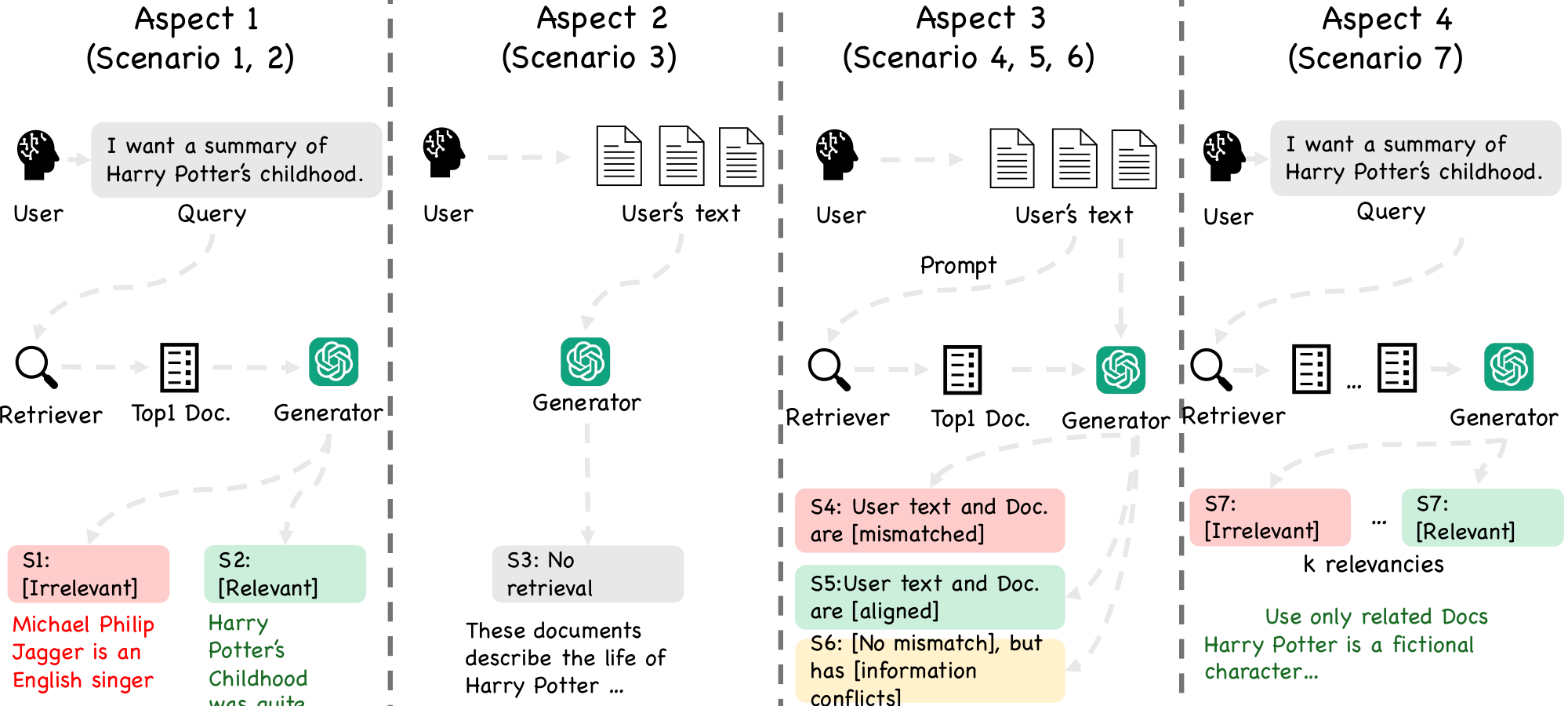}
\caption{LogicSumm's \cgh{pipeline, which} divides evaluation into four aspects and seven scenarios}
\label{fig:LogicSumm}
\vspace*{-1em}
\end{figure}

\cgh{\textit{LogicSumm} builds a structured foundation for testing by defining seven common summarization scenarios divided into four higher-level aspects.} We \cgh{begin by presenting} the overall pipeline for \cgh{evaluating a} summarization task. \cgh{Our} framework is depicted in Figure~\ref{fig:LogicSumm}. 

\textbf{Problem Formulation.} For a given user query $\mathbf{q}$, the retriever $\mathbf{R}$ is tasked with fetching the top-$\mathbf{k}$ documents $\{\mathbf{D}_1, \dots, \mathbf{D}_k\}, \mathbf{k} \geq 1$ from a database of document vectors $\mathbf{D}$ \cgh{via} a semantic similarity search mechanism. The LLMs' generator $\mathbf{G}$ produces a summary based on different source information. We formally define four aspects, where an \textit{aspect} is a high-level query type that $\mathbf{R}$ needs to answer, and a \textit{scenario} is a particular type of sub-aspect with unique scenario properties.

\begin{align}
\text{Aspect 1: } & \qquad \textit{Summarization} = \mathbf{G}(\mathbf{R}_{top1}(\mathbf{q})), \nonumber \\
\text{Aspect 2: } & \qquad \textit{Summarization} = \mathbf{G}(\mathbf{q}), \nonumber \\
\text{Aspect 3: } & \qquad \textit{Summarization} = \mathbf{G}(\mathbf{q} \oplus \mathbf{R}_{top1}(\mathbf{q})), \nonumber \\
\text{Aspect 4: } & \qquad \textit{Summarization} = \mathbf{G}(\mathbf{R}_{topk}(\mathbf{q})), \nonumber
\end{align}

 where $\mathbf{R}(\mathbf{q}) = \{\mathbf{D}_1, \dots, \mathbf{D}_k\}$ and $\oplus$ is the string concatenation \cgh{operator}. Summarization quality is \cgh{heavily} influenced by the accuracy of the retriever $\mathbf{R}$ and the quality of the document vector store $\mathbf{D}$ from which information is sourced. 

\textbf{\cgh{Aspect Scenarios.}} In each constructed scenario, we expect that LLMs will not only undertake actions with logical precision but also exhibit \cgh{high-quality} summarization capabilities. 

\begin{itemize}

\item \textbf{Aspect 1: Scenarios $1$, $2$.} \cgh{The} LLMs \cgh{should} discern the relevance of $\mathbf{D}_1$ to query $\mathbf{q}$. 

\item \textbf{Aspect 2: Scenario $3$.} The LLMs \cgh{must} summarize the user's provided text directly. 

\item \textbf{Aspect 3: Scenario $4$.} The LLMs \cgh{are expected} to indicate the \cgh{lack of relevance} of $\mathbf{D}_1$ to the user's text and suggest summarizing solely based on the user's text.

\item \textbf{Aspect 3: Scenario $5$.} \cgh{The} LLMs should recognize \cgh{both} the relevance and \cgh{the} absence of conflict between the user's text and $\mathbf{D}_1$, \cgh{then} summarize both sources. 

\item \textbf{Aspect 3: Scenario $6$.} The LLMs \cgh{must} identify relevance coupled with an informational conflict between the user's text and $\mathbf{D}_1$.

\item \textbf{Aspect 4: Scenario $7$.} \cgh{The} LLMs are expected to \cgh{recognize} the relevance of $\mathbf{R(q)}$ to the user's query $\mathbf{q}$ and exclude any irrelevant documents from the summarization.

\end{itemize}

\textbf{Motivating Observations:} With the introduction of \textit{LogicSumm} we are equipped to assess the \cgh{real-world} proficiency of LLMs \cgh{leveraging RAGs} to \cgh{perform} summarization tasks. We deploy the Mistral-7B Instruct model \cite{jiang2023mistral} for this evaluation. 

\subsection{Implementation and Evaluation Metrics}

 Our evaluation establishes baselines using a collection of autoregressive LLMs based on GPT: GPT-3.5, Claude 2, Jurassic, and LLaMa2-13B, conducted in a zero-shot manner where instructions were given to complete tasks within the \textit{LogicSumm} framework. \cgh{We also applied advanced prompting techniques for the Mistral-7B Instruct model in both zero-shot and one-shot ``Chain of Thought'' contexts \cite{wei2023chainofthought}}. Within each aspect, the same prompt is used for all scenarios it includes. The prompt details are \cgh{described} in Appendix \ref{app: eval prompts}.
 
 Our assessment criteria included not only \cgh{\textit{LogicSumm}'s} logical accuracy but also the quality of the summaries evaluated using BertScore \cite{zhang2020bertscore} and Rouge 1/2/L \cite{lin-2004-rouge}. We employ GPT-4 Turbo to assess whether a model's output maintains logical correctness. \cgh{It is} important to note that summary quality was assessed only in Scenarios 2 and 3, as the text retrieval in other scenarios was deemed irrelevant to the user's text, rendering summary quality evaluation unnecessary. Evaluating logical accuracy in Scenario 3 is also unnecessary because it is \cgh{a} direct summarization of the user's text. We follow \cgh{a} procedure to generate test data \cgh{similar to} the method employed \cgh{to create} training data. The gold summaries are derived from the outputs produced by GPT-4 Turbo. \cgh{For the seven scenarios we generated 57, 48, 50, 36, 50, 43, and 98 samples, respectively.}

%

\begin{figure}[!t] 
\centering 
 \includegraphics[width=1\linewidth]{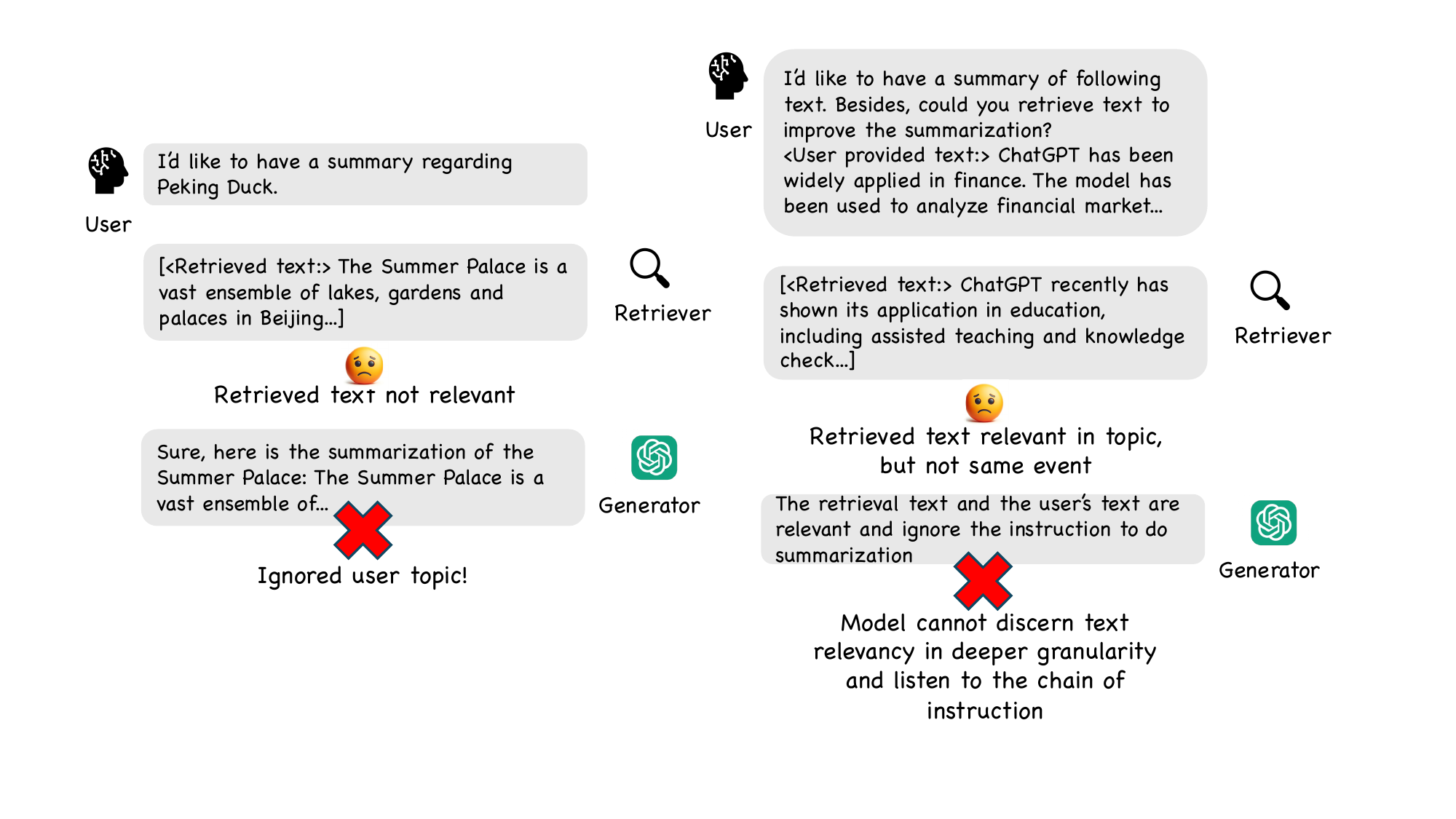}
 \vspace*{-0.5in}
 \caption{Illustration of \cgh{limitations} under \cgh{\textit{LogicSumm}}}
 \label{fig:limit1}
 \vspace*{-1em}
\end{figure}

To examine performance with multiple top-ranked documents where \(k > 1\), we evaluated summary quality for \(k = 5, 8,\) and \(10\), simulating situations with five relevant documents. In instances where \(k = 8\) and \(k = 10\) we introduced three and five irrelevant documents to test our model's resilience in handling irrelevant content within the top-ranked documents. We benchmark our method against other general RAG-based summarization frameworks \cgh{including} \textit{Stuff Summarization}, \textit{Map-Reduce Summarization}, and \textit{Refine Summarization} utilizing Mistral-7B Instruct. Additionally, we provide explict instructions to disregard irrelevant documents. The specifics of the prompts can be found in Appendix \ref{app: multi}.

Our observations \cgh{suggest} that \textit{LogicSumm} exhibits limitations \cgh{when attempting to recognize} the relevance between \cgh{a} user's text \cgh{while} following a sequence of instructions: first assessing the relevance of the retrieval text, then determining whether to proceed with summarization. \cgh{Testing} prompts can be found in Appendix \ref{app: me}. \cgh{Detailed explanations are included in the Experiment Results section below. We defer this discussion until introducing \textit{SummRAG}, since \textit{SummRAG} performance is compared to existing, state-of-the-art approaches.}

\section{SummRAG}
\label{sec: summrag}

Initial findings from \textit{LogicSumm} suggest that general-purpose \cgh{LLMs} may not be \cgh{sufficiently} robust for RAG-based summarization. This led to a complete system \textit{SummRAG} that \cgh{creates and fine-tunes} dialogues \cgh{and models} with GPT-4 Turbo to \cgh{produce more reliable LLMs for each} situation \cgh{tested with} \textit{LogicSumm}. We \cgh{begin} by creating special tokens embedded in the generated dialogue to ensure it has \cgh{a proper} format. We then focus on the top-$\mathbf{1}$ document case (Aspects $1, 2$, and $3$) before moving to situations involving the top-$\mathbf{k}$ documents, \(k > 1\) (Aspect 4.) \cgh{We conclude by} fine-tuning the model using the dialogues we produce.

\begin{table}[!t]
\caption{Description of special tokens}
\label{tab:spec_tok}
\footnotesize
\begin{tblr}{
  colspec={Q[h] X[l,b] Q[h,c]},
  row{even}={bg=azure9},
  row{odd}={bg=gray9},
  row{1}={c,bg=gray4,fg=white},
}
Type & Definitions & Aspect \\
{[Retrieval]}, [No Retrieval] & Retrieval needed & 1--4\\
{[Retrieval]}, [Irrelevant] & Retrieval text is relevant to the user's text & 1--4 \\
{[Continue to use User's Text]} & Retrieval text is not relevant to the user's text & 3 \\
{[Information Conflict}] & Retrieval text is relevant to the user's text but there is an information conflict between them & 3 \\
{[Augmenting User's Text}] & Retrieval text is relevant to the user's text with no information conflict & 3 \\
{[Context]}, [/Context] & An intermediate summarization & 4 \\
$<$Count$>$, $<$/Count$>$ & Count documents left to summarize & 4 \\
\SetRow{gray9} {[Topic]} & Memorize the user's topic & 4 \\
\end{tblr}
\vspace*{-1em}
\end{table}

\cgh{Throughout} this section we use $\textbf{D}$ to represent the collection of document vectors, \cgh{including} datasets from CNN Daily Mail and XSum available in the HuggingFace repository. $\textbf{R}$ \cgh{is} the retriever used in our framework, \cgh{and} $\textbf{t}$ is the user's text. Additionally, $\mathbf{D_s} = \{D_1, \ldots, D_k\}$ \cgh{denotes} the collection of retrieved documents ranked based on their semantic similarity. $\mathbf{D_r}$ \cgh{is} a random document selected from $\textbf{D}$.

\subsection{Logical Special Tokens}

We \cgh{insert} logical special tokens whose meanings are introduced to GPT-4 during dialogue creation, \cgh{providing} clarity and compactness in the conversation while ensuring proper formatting. \cgh{In} the later stages of model fine-tuning we substitute these tokens with natural language text \cgh{to avoid the} extensive instruction data \cgh{required to extend the LLMs to automatically manage the special tokens.}

%

In addition to the tokens outlined in Table~\ref{tab:spec_tok}, we incorporate function-calling tokens $\textbf{[API]}, \textbf{[/API]}, \textbf{[Argument]}$ within the generated dialogue \cite{qin2023toolllm}. This allows \cgh{the} LLMs to interface with our custom text mining APIs\footnote{\url{https://go.ncsu.edu/social-media-viz}} \cgh{that} are capable of performing tasks such as analyzing the sentiment of summaries or accessing online news sources to generate insightful sentiment visualizations. 

\subsection{Dialogue Generation: Top-\texorpdfstring{$\mathbf{1}$}{1} Document}

The top-$\mathbf{1}$ document scenario encompasses Aspects 1, 2, and 3 in \textit{LogiSumm}. We \cgh{apply} GPT-4 Turbo to generate the dialogue by introducing the meaning of the special tokens and providing a one-shot demonstration in the prompts. 

\begin{figure}[!t] 
\centering 
 \includegraphics[width=1\linewidth]{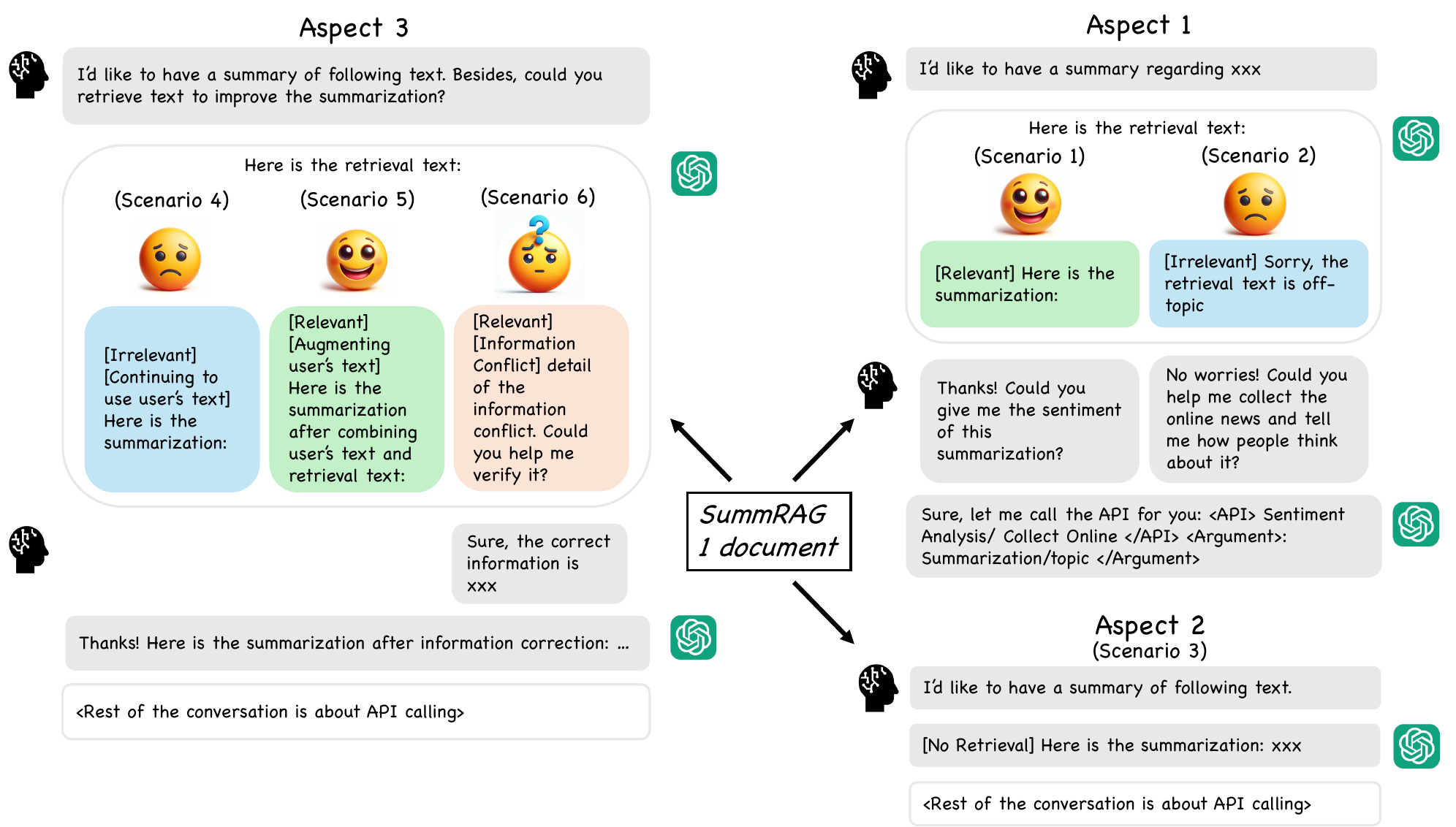}
\caption{Dialogue generation for the top-$\mathbf{1}$ document}
\label{fig:SummRAG1}
\vspace*{-1em}
\end{figure}

\begin{figure}[!t] 
\centering 
\includegraphics[width=0.9\linewidth]{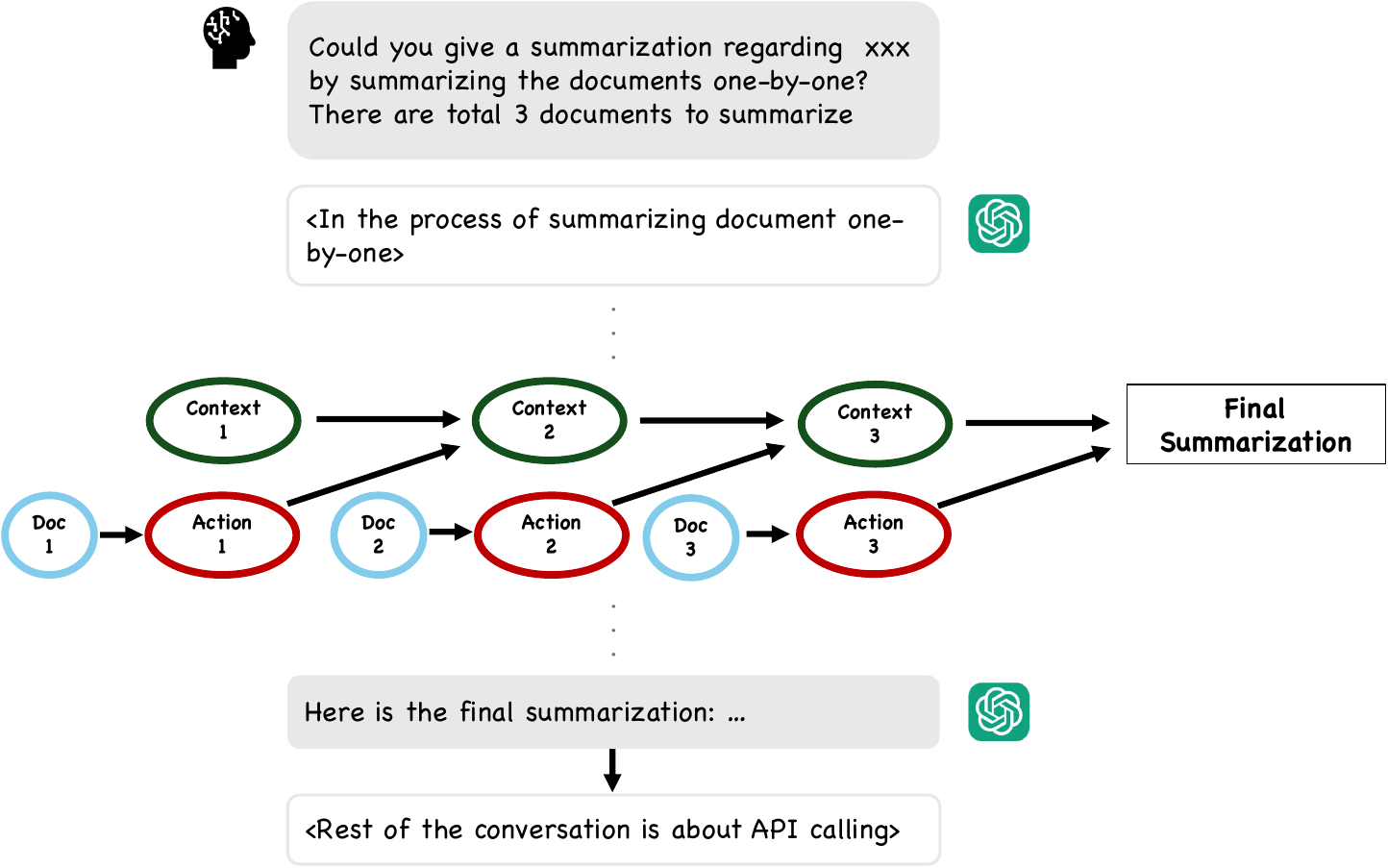}
\caption{Dialogue generation for the top-$\mathbf{k}$ documents}
\label{fig:SummRAG2}
\vspace*{-1em}
\end{figure}

The utility of the special tokens defined in Table~\ref{tab:spec_tok} is illustrated in Figure \ref{fig:SummRAG1}. To improve diversity in conversations, we \cgh{instruct} GPT-4 Turbo to incorporate variations for certain sentences, such as \textit{``I'd like to have a summary xxx''}, in the user instruction \cgh{component}. For Aspect 1, Scenario \cgh{1} is based on a random document $\mathbf{D_r}$ and its topic \textbf{t} extracted using GPT-3.5 Turbo for relevancy. Scenario 2 involves a randomly chosen unrelated topic to create irrelevancy. For Aspect 2, Scenario 3 uses $\mathbf{D_r}$ to represent the user's text. For Aspect 3, Scenario 4 utilizes two random documents $\mathbf{D_{r,1}}$ and $\mathbf{D_{r,2}}$ to introduce irrelevancy. In Scenario 5, GPT-4 Turbo is prompted to generate topics it can output \cgh{as} factual stories, then to create two documents on the same subtopic to ensure relevancy. Scenario 6 \cgh{instructs} GPT-4 Turbo to introduce information conflicts in $\mathbf{D_r}$, such as changes in numbers, factual reversals, and date alterations. This pair of documents, showcasing information conflict, represents the user's input and the retrieved text, respectively. 

The meanings of the special tokens, the one-shot demonstration, \cgh{and} the documents for each aspect \cgh{described} above are provided as prompts to GPT-4 Turbo to generate the intended dialogue. Details of these prompts are available in the Appendix \ref{app: prompts}.

\subsection{Dialogue Generation: Top-\texorpdfstring{$\mathbf{k}$}{k} \cgh{Documents}}

\begin{wrapfigure}{l}{0.6\linewidth}
\centering
\includegraphics[width=\linewidth]{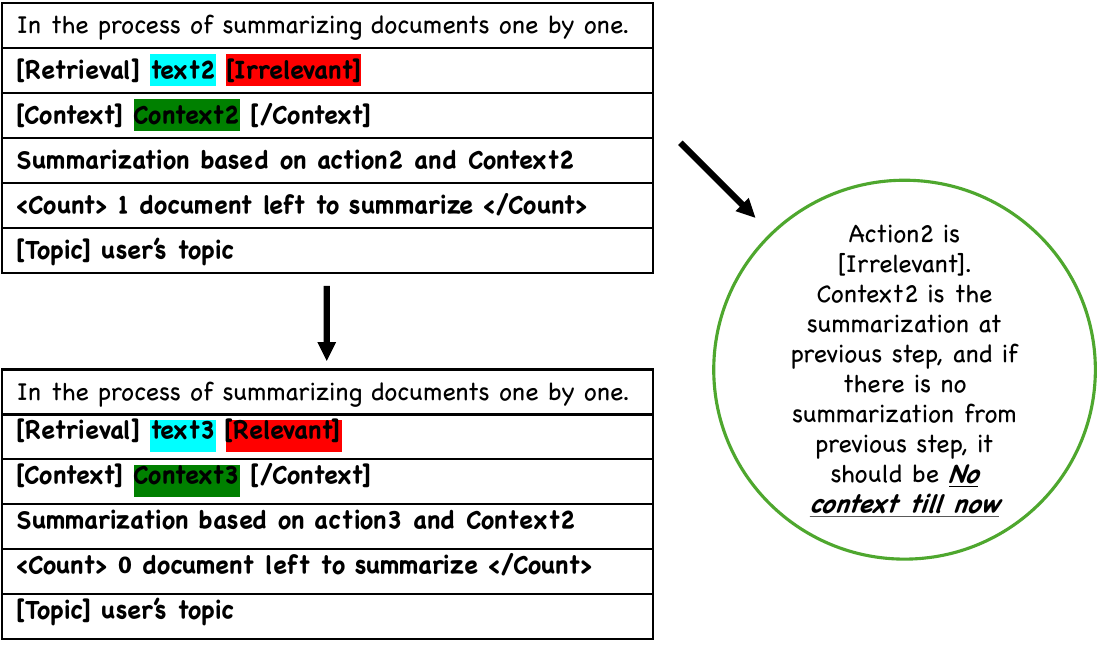}
\caption{Example dialogue at each \cgh{summarization} step}
\label{fig:SummRAG3}
\end{wrapfigure}

To transition from the top-$\mathbf{1}$ document to the top-$\mathbf{k}$ documents (Aspect 4) we introduce the notion of \textbf{context} $\mathit{ctx}$. This concept represents a text segment that \cgh{stores} the intermediate state of multi-document summarization. It \cgh{allows} LLMs to adopt a Markov-like thought process for summarizing documents, \cgh{where} the summarization at each step relies solely on $\mathit{ctx}$ and the document retrieved at that particular step (see Figures \ref{fig:SummRAG2},~\ref{fig:SummRAG3}). This \cgh{frees} LLMs from storing all the documents in the input prompt. 

The special tokens $<$Count$>$ \textit{0 documents left to summarize} $<$/Count$>$ serve as \cgh{a} stopping \cgh{criteria} indicating there are no more documents to summarize. \cgh{At this point} the LLM returns the final summarization to the user. We generate conversations for the top-$\mathbf{5}$ scenario where $\mathbf{D_s} = \{D_1, D_2, D_3, D_4, D_5\}$. However, our experiments demonstrate that utilizing chat-based models with general instruction capabilities such as Mistral-7B Instruct \cgh{does not} limit the multi-document summarization to \cgh{only five} documents. This is achieved by strategically using the $<$Count$>$ token \cgh{to allow} flexibility in the number of summarized documents. The step-by-step prompts \cgh{are} available in Appendix \ref{app: prompts}.

\subsection{Model \cgh{Fine-Tuning}}
 
\cgh{We begin by collecting} a custom dialogue dataset \cgh{to use} to fine-tune the Mistral-7B Instruct model checkpoints. However, we encountered difficulties in teaching the model to understand the special token definitions. To address this, we convert \cgh{the} tokens into text using a transformation table (Appendix~\ref{app:transform_text}). \cgh{We also insert} aspect-specific system prefixes to further guide the model's learning process. Function-calling tokens $\textbf{[API]}, \textbf{[/API]}$, and $\textbf{[Argument]}$ within the generated dialogue are changed to the text ``\textit{Here is the API: }'' and ``\textit{The argument of the API: }''.

\cgh{Next, we collect training data to fine-tune the Mistral-7B Instruct model}. For Aspects 1, 2, and 3 we \cgh{apply} the chat template $\langle s \rangle [INST]\ \dots\ [/INST]\ \dots \ \langle /s \rangle$. For Aspect 4, rather than training on the entire dialogue, we focus on adjacent pairs of steps. \cgh{Here}, the previous step serves as the instruction and the subsequent step as the response. \cgh{The} retrieval text is masked to ensure it conforms to the correct format. 
\cgh{Given} a dialogue $d \sim \mathcal{D}_{gen}$ from the \cgh{custom} dataset where $\mathcal{D}_{gen}$ is the data distribution implicitly defined in the dialogue generation process, we train our model $\mathcal{M}_{\theta}$ on $\mathcal{D}_{gen}$ using the standard next token objective: 
\begin{equation}
\max_{\theta} \mathbb{E}_{d \sim \mathcal{D}_{gen}}\log p_{\mathcal{M}_{\theta}}(y | x)
\end{equation}
where $x$ is the instruction and retrieval text \cgh{within} $d$ and $y$ is the response. We use LoRA (\cite{hu2021lora} to perform parameter-efficient tuning and store adapter weights.

\subsection{Connection to Prior Work}

\textit{SummRAG} \cgh{modifies and expands on} \textit{Self-RAG} \cite{asai2023selfrag} to \cgh{address} the specific needs of summarization in the context of RAG, specifically:

\begin{enumerate}

\item Shift the granularity of the critical thinking process from individual sentences to the full retrieval text.
    
\item Utilize special tokens during dialogue generation with GPT-4 Turbo, then replace tokens with natural language expressions during fine-tuning of the model.

\end{enumerate}

Rather than exploring question-answering \textit{knowledge conflicts} \cite{jin2024tug}, our research concentrates on summarization, employing a comprehensive evaluation pipeline. We also create a curated dataset to address the challenges outlined in our evaluation framework.

\section{Experiments}

To conduct \cgh{our} evaluation, we establish baselines using autoregressive LLMs GPT-3.5, Claude 2, Jurassic, and LLaMa2-13B in a zero-shot manner where instructions complete tasks within the \textit{LogicSumm} framework. \cgh{We then applied advanced prompting techniques with} the Mistral-7B Instruct model in both zero-shot and one-shot ``Chain of Thought'' contexts. Within each aspect, the same prompt \cgh{was} used for all scenarios (Appendix \ref{app: eval prompts}).

\subsection{Implementation and Evaluation Metrics}

%

\begin{table}[!t]
\caption{Comparison of different methods with respect to logical accuracy under \textit{LogigSumm}, the best results are shown in \textbf{bold} with reported results based on the test dataset}
\label{tab:res_logi_acc}
\centering
\footnotesize
\begin{tblr}{
  colspec={l c c c c c c},
  row{odd}={bg=azure9},
  row{even}={bg=gray9},
  row{1-2}={c,bg=gray4,fg=white},
  row{7}={abovesep=4pt}
}
 & \SetCell[c=6]{c} Scenario\\
LLM & 1 & 2 & 4 & 5 & 6 & 7\\
Claude2 & 0.96 & \textbf{1.0} & 0.88 & \textbf{1.0} & \textbf{0.60} & --\\
Jurassic & 0.98 & 0.58 & 0.84 & \textbf{1.0} & 0.26 & --\\
Llama2 13B Chat & 0.88 & \textbf{1.0} & 0.84 & 0.86 & 0.56 & --\\
GPT-3.5 Turbo & 0.96 & \textbf{1.0} & \textbf{1.0} & \textbf{1.0} & 0.52 & --\\
\hline[dotted]
Minstral 7B Chat (explicit logical instructions) & 0.29 & \textbf{1.0} & \textbf{1.0} & 0.14 & 0.58 & --\\
Minstral 7B Chat (zero-shot Chain of Thought) & 0.88 & \textbf{1.0} & 0.97 & 0.80 & -- & --\\
Minstral 7B Chat (one-shot Chain of Thought) & \textbf{1.0} & 0.19 & 0.91 & 0.88 & -- & --\\
\hline[dotted]
SummRAG & \textbf{1.0} & \textbf{1.0} & 0.97 & \textbf{1.0} & \textbf{0.79} & \textbf{0.86} \\ 
\end{tblr}
\end{table}

\begin{table}[!t]
\caption{Average summarization quality in Scenarios 2 and 3, the best results are in \textbf{bold}}
\label{tab:avg_summ_q}
\centering
\footnotesize
\begin{tblr}{
  colspec={l c c c c},
  row{odd}={bg=azure9},
  row{even}={bg=gray9},
  row{1-2}={c,bg=gray4,fg=white},
}
 & \SetCell[c=2]{c}{Scenario 2} & & \SetCell[c=2]{c}{Scenario 3} &  \\
LLM & {BertScore\\precision, recall, F1} & {Rogue\\1, 2, L} & {BertScore\\precision, recall, F1} & {Rogue\\1, 2, L}\\
GPT-3.5 Turbo & $\mathbf{0.91},\mathbf{0.91},0.90$ & $\mathbf{0.48},\mathbf{0.21},\mathbf{0.33}$ & $\mathbf{0.91},\mathbf{0.91},\mathbf{0.91}$ & $\mathbf{0.50},0.24,\mathbf{0.37}$\\
Llama 13B Chat & $0.89,0.90,0.89$ & $0.42,0.18,0.30$ & $0.90,\mathbf{0.91},0.90$ & $0.45,0.19,0.31$\\
Mistral-7B Instruct & $0.90,0.90,0.90$ & $0.44,0.19,0.31$ & $0.90,0.90,0.90$ & $0.42,\mathbf{0.28},0.31$\\
\hline[dotted]
SummRAG & $\mathbf{0.91},0.90,\mathbf{0.91}$ & $\mathbf{0.48},\mathbf{0.21},\mathbf{0.33}$ & $\mathbf{0.91},0.90,\mathbf{0.91}$ & $0.48,0.22,0.35$  \\ \hline
\end{tblr}
\vspace*{-1em}
\end{table}

Our assessment criteria \cgh{includes} not only logical accuracy within the \textit{LogicSumm} context but also the quality of the summaries evaluated using BertScore and Rouge 1, 2, L. GPT-4 Turbo assesses the model's logical correctness. \cgh{Note that} summary quality was assessed only in Scenarios 2 and 3, as the text retrieval in \cgh{the} other scenarios was deemed irrelevant to the user's text, rendering summary quality evaluation unnecessary. Evaluating logical accuracy in Scenario 3 is also \cgh{irrelevant} because it is \cgh{a} direct summarization of the user's text. We follow the same procedure to generate test data \cgh{that was used to create} training data. The gold standard summaries are derived from the outputs produced by GPT-4 Turbo.

\cgh{To} examine our model's performance with multiple top-ranked documents where \(k > 1\) we evaluated summary quality for \(k = 5, 8,\) and \(10\). For $k=5$ we generated five relevant documents. For \(k = 8\) and \(k = 10\) we introduced three and five irrelevant documents, respectively. We \cgh{compared} our method against other RAG-based summarization frameworks \cgh{including} \textit{Stuff Summarization}, \textit{Map-Reduce Summarization}, and \textit{Refine Summarization}
utilizing Mistral-7B Instruct as the LLM engine within these frameworks. Additionally, we \cgh{provided} explicit instructions to disregard irrelevant documents. The prompts are in Appendix \ref{app: multi}.

\subsection{Results}

\cgh{Results} from \cgh{the} \textit{LogicSumm} scenarios led to two key findings. \cgh{First,} the logical accuracy of Mistral-7B Chat (Table~\ref{tab:res_logi_acc}) \cgh{varies significantly} based on the selected prompts. When explicit logical instructions guide Mistral-7B Chat, it demonstrates lower accuracy in Scenarios 1 and 5 and higher accuracy in Scenario 2, but with explicit guidance and one-shot Chain of Thought \cgh{it} returns lower accuracy in Scenario 2 and higher accuracy in Scenarios 1 and 5. In Scenario 6 the Chain of Thought prompting strategy \cgh{struggles} to \cgh{identify} information conflicts. This indicates that devising a prompting strategy that consistently maintains robust \cgh{performance} across different scenarios can be time-consuming and challenging.


\cgh{Second}, after fine-tuning Mistral-7B Chat on our curated training dataset, the logical accuracy across all aspects remains consistently high compared to other models. This underscores the effectiveness of \textit{SummRAG}. \textit{SummRAG} creates data without adding new knowledge to the model. This implies that the model possesses sufficient understanding of the logic required but benefits from instruction-tuning to guide its application of this knowledge.

%

\textit{SummRAG} enhances robustness \cgh{while maintaining} the quality of summarization, as \cgh{shown} in Table~\ref{tab:avg_summ_q}. It \cgh{produces results comparable to} GPT-3.5 Turbo and slightly outperforms Llama 13B Chat and Mistral-7B Chat. Given that GPT-4 Turbo's outputs serve as the gold standard for summarization, this indicates that during the instruction tuning phase \textit{SummRAG} enables Mistral-7B Chat to \cgh{match} GPT-4 Turbo's summarization capabilities.

In the multi-document setting, Table~\ref{tab:summ_perf} demonstrates that as the count of irrelevant documents increases the performance metrics tend to decrease across other summarization frameworks. This indicates that simply using prompts to disregard irrelevant documents may not be a robust approach. In contrast, internalizing the concept of context \cgh{demonstrates} resilience against the presence of irrelevant documents (\textit{e.g.}, Scenario 7). This is achieved without significantly increasing inference costs, as the model at each inference step depends solely on $\mathit{ctx}$ and the text retrieved at that particular step.

\begin{table}[!t]
\caption{Summarization performance across different document sets, the best results are in \textbf{bold}}
\label{tab:summ_perf}
\centering
\footnotesize
\begin{tblr}{
  colspec={l c c c c c c},
  colsep={1.5pt},
  row{4,6}={bg=azure9},
  row{5,7}={bg=gray9},
  row{1-3}={c,bg=gray4,fg=white},
}
\SetCell[r=3]{c} Format & \SetCell[c=6]{c} Summarization Score\\
 & \SetCell[c=2]{c} 5 Documents & & \SetCell[c=2]{c} 8 Documents & & \SetCell[c=2]{c} 10 Documents\\
 & {BertScore\\P, R, F1} & {Rogue\\1, 2, L} & {BertScore\\P, R, F1} & {Rogue\\1, 2, L} & {BertScore\\P, R, F1} & {Rogue\\1, 2, L}\\
{\fontsize{7}{8}\selectfont Stuff} & $0.85, \mathbf{0.88}, 0.87$ & $\mathbf{0.40},\mathbf{0.16}, 0.23$ & $0.85, \mathbf{0.88}, 0.86$ & $0.39, \mathbf{0.16}, 0.21$ & $0.84, 0.86, 0.85$ & $0.35, 0.12, 0.19$\\
{\fontsize{7}{8}\selectfont Map-Reduce} & $0.85, 0.87, 0.86$ & $0.38, 0.13, 0.21$ & $0.83, 0.86, 0.84$ & $0.32, 0.11, 0.18$ & $0.82, 0.85, 0.84$ & $0.31, 0.09, 0.17$\\
{\fontsize{7}{8}\selectfont Refine} & $0.86, 0.85, 0.85$ & $0.30, 0.08, 0.17$ & $0.83, 0.82, 0.82$ & $0.19, 0.12, 0.12$ & $0.83, 0.83, 0.83$ & $0.24, 0.04, 0.15$\\
{\fontsize{7}{8}\selectfont SummRAG} & $\mathbf{0.88}, \mathbf{0.88}, \mathbf{0.88}$ & $\mathbf{0.40}, 0.15, \mathbf{0.24}$ & $\mathbf{0.87}, 0.87, \mathbf{0.87}$ & $\mathbf{0.41}, 0.13, \mathbf{0.22}$ & $\mathbf{0.87}, \mathbf{0.87}, \mathbf{0.87}$ & $\mathbf{0.40}, \mathbf{0.14}, \mathbf{0.24}$\\
\end{tblr}
\vspace*{-2em}
\end{table}

\subsection{Supporting Analysis}

To \cgh{better understand} the logical reasoning capabilities of an LLM trained with \textit{SummRAG}, we \cgh{examine} distribution shift from the original to the fine-tuned model. One \cgh{case} involves a scenario with a user query: ``\textit{I'd like a summary about the American stock market}'' and the text retrieved relates to Amazon's stock information. The original model generates a summary based solely on Amazon stock \cgh{details}. In contrast, after being fine-tuned with \textit{SummRAG}, our model responds by indicating, ``\textit{The retrieved text does not offer insights into the overall performance of the American stock market but instead concentrates on future projections for Amazon's stock price}.'' If the user's query were \cgh{changed} to: ``\textit{I'd like a summary about different companies' performances in the American stock market}'' our fine-tuned model would recognize the relevance of the Amazon stock information to this broader query and proceed to summarize the text accordingly. Our fine-tuned model possesses a more nuanced understanding of relevance, assessing the content with greater depth and specificity. 

In practical applications when segmenting text \cgh{it is} common to use overlapping chunks to prevent \cgh{discontinuities} in information flow. Notably, our model remains robust even when there \cgh{is} unrelated material in the retrieval text. \cgh{Our model does not} mistakenly \cgh{consider} an \cgh{entire} segment irrelevant due to a small amount of unrelated content.

\section{Conclusion, Limitation, and Future Work}

In this paper, we propose a new evaluation framework \textit{LogicSumm} designed to assess the robustness of LLMs within the context of RAG-based summarization. \cgh{Based on limitations identified with \textit{LogiSumm}}, we \cgh{developed} \textit{SummRAG}, a comprehensive system that spans generating training dialogues to fine-tuning LLMs. \cgh{\textit{SummRAG} is designed to enhance} robustness in \textit{LogicSumm}'s scenarios. Experiments focusing on logical accuracy and \cgh{sumnmarization} quality \cgh{confirm} the effectiveness of \textit{SummRAG}. \cgh{Further improvements must still be considered, however}. \textit{SummRAG}'s performance is linked to the scenarios in \textit{LogicSumm}, which may not encompass all possible real-life situations. \cgh{This suggests} a need for a more inclusive evaluation framework. Furthermore, the efficacy of our approach is influenced by the quality of the prompts used \cgh{during} dialogue generation, highlighting the potential advantages of developing a more automated strategy for prompt selection in future work.

\bibliography{colm2024_conference}
\bibliographystyle{colm2024_conference}

\appendix
\section{Appendix}

\subsection{Motivating Examples and Testing Prompts}
\label{app: me}
We test FlanT5-xxl\cite{chung2022scaling} and Llama2 13B Chat with following prompt: 

------------------------------------- 

\textit{\textbf{User}}: 

\textbf{I'd like a summary regarding Peking Duck. Here is the retrieval text}: [ChatGPT is similar to the automated chat services found on customer service websites, as people can ask it questions or request clarification to ChatGPT's replies. The GPT stands for "Generative Pre-trained Transformer," which refers to how ChatGPT processes requests and formulates responses. ]\textbf{Please give me the summary based on the retrieval text}. 

\textit{\textbf{FlanT5-xxl}}: 

ChatGPT is similar to the automated chat services found on customer service websites, as people can ask it questions or request clarification to ChatGPT’s replies. The GPT stands for ”Generative Pre-trained Transformer,” which refers to how ChatGPT processes requests and formulates responses. 

------------------------------------- 

\textit{\textbf{User}}: 

\textbf{I'd like a summary regarding ChatGPT application in Finance. Here is the retrieval text}: [It’s easy to understand why educators feel threatened. ChatGPT is a freakishly capable tool that landed in their midst with no warning, and it performs reasonably well across a wide variety of tasks and academic subjects]\textbf{Please give me the summary based on the retrieval text}. 

\textit{\textbf{Llama2 13B Chat}}: 

Certainly! Based on the retrieval text, here is a summary about the potential applications of ChatGPT in finance: .... 

------------------------------------- 

The examples provided demonstrate that these two LLMs exhibit a lack of robustness in the scenarios we have presented.

\subsection{Prompts to Generate Dialogue}Below, we show how we generate the dialogue for Scenario $1$, $4$, and $7$ (Prompts for other scenarios can be found in our code repository) 
\label{app: prompts} 

\textit{\textbf{Scenario 1}}: 

\textbf{Here is the one-shot example to show GPT-4 Turbo}: 

[{"role": "user", "content": "I need a summary about the incident involving Chen Xing."}, {"role": "assistant", "content": "[Retrieval]$\langle$paragraph$\rangle$ OpenAI's ChatGPT is an AI-powered chatbot with natural language processing capabilities, designed to engage in human-like dialogues. As a generative AI model, it employs unsupervised learning algorithms to produce textual, code, and digital image outputs. Despite not being the sole AI chatbot available, ChatGPT's rapid content generation abilities have contributed significantly to its popularity. Utilizing appropriate prompts with ChatGPT facilitates the acquisition of requisite information. $\langle /$paragraph$\rangle$[Irrelevant] Unfortunately, the retrieved information does not pertain to the user's inquiry."}, {"role": "user", "content": "Ok. If there is no related document in the database, could you collect the online comments and show me what people think now? Thanks!"}, {"role": "assistant", "content": "No problem, I will initiate the API request for you. $\langle$API$\rangle$Collect{\_}Online$\langle /$API$\rangle$[Argument] Chen Xing[/Argument]"}] 

\textbf{Here is the instruction to tell GPT-4 Turbo}: 

You are instructed to construct the conversation between the assistant and the user where the user requires the assistant to do summarization on some topic and some post-summarization analysis. Some special tokens need to be added to the conversation. You are required to follow the format of the provided example, including the position of special tokens. Here are special tokens: 

[1] [No Retrieval] and [Retrieval] are used to decide whether you need to retrieve the text for summarization. 

[2]  [Irrelevant] and [Relevant] are used to check whether the retrieval text is related to the user query. 

The retrieval text will be inside of $\langle$paragraph$\rangle$ and $\langle /$paragraph $\rangle$ and you need to remove $"$ in the retrieval text. The summary will be appended after the [Relevant] or [Irrelevant]. Besides, there is one API called Collect$\_$Online. It is used when the user wants to collect online news and know people's attitude. The argument of this API is the topic. 

Here is one example: 

${\#\#\#}$ \textbf{one-shot example} ${\#\#\#}$ 

Now, I will only provide you with the topic and the retrieval text. You decide the rest of them and construct the conversation following STRICTLY the same format as the provided example.
Note that the user may collect the online news. User will not say the name of the API in the query since they don't know the existence of API. You need to always use [Retrieval] in this case. 

Here is the topic: \textbf{user topic}

Here is the retrieval text:  \textbf{retrieval text}  

\textit{\textbf{Scenario 4}}: 

\textbf{Here is the one-shot example to show GPT-4 Turbo:} 

[{"role": "user", "content": "Hi, could you summarize the following text for me? Besides, could you also retrieve some related text and see if it can improve the summarization and also check the information conflict"}, {"role": "assistant", "content": "Sure, could you provide the text? "}, {"role": "user", "content": "\textbf{user's own text}"}, {"role": "assistant", "content": "[Retrieval] \textbf{Irrelevant retrieval text}[Irrelevant][Continuing to use User's text]The summarization of user's text"}] 

\textbf{Here is the instruction to tell GPT-4 Turbo}: 

You are instructed to construct the conversation between the assistant and the user where the user requires the assistant to do summarization on some topic and some post-summarization analysis. Some special tokens need to be added to the conversation. You are required to follow the format of the provided example, including the position of special tokens. Here are special tokens: 

        [1] [No Retrieval] and [Retrieval] are used to decide whether you need to retrieve the text for summarization. 
        
        [2] [Irrelevant] and [Relevant] are used to check whether the retrieval text is related to the user query. 
        
        [3] [Continuing to use User's text] is used to when the retrieval text is not related to user query and appended after [Irrelevant]. 
        
        The retrieval text will be inside of $\langle$ paragraph$\rangle$ and $\langle /$paragraph $\rangle$ and you need to remove $"$ in the retrieval text. The summary will be appended after [Continuing to use User's text] . Besides, there are total four APIs: 
        Sentiment$\_$Analysis, Collect$\_$Online, Knowledge$\_$Construction. 
        
        [1] Sentiment$\_$Analysis is used when user wants to know the sentiemnt of the summary. The argument is the summary. 
        
        [2] Collect$\_$Online is used when user wants to collect online news and know people's attitude. The argument is the topic 
        
        [3] Knowledge$\_$Construction is used when user wants to construct the knowledge graph based on the summary. The argument is the summary. 
        
        Here is one example: 

        ${\#\#\#}$ \textbf{one-shot example} ${\#\#\#}$ 

        Now, I will only provide you the user's text and the retrieval text and you need to remove " in the user text. You decide the rest of them and construct DIVERSE conversation following the same format as the provided example.
       . [Continuing to use User's text]  MAKE SURE to IGNORE the RETRIEVAL TEXT when doing summarization. You need to always use [Retrieval] in this case
        You can use one, two or three APIs. 
        
        Here is the user's text: 
        \textbf{user's own text} 
        
        Here is the retrieval text: 
        \textbf{retrieval text} 

\textit{\textbf{Scenario 7}}: 

It should be noted that the conversation for Scenario 7 is structured as a step-by-step construction due to the observation that GPT-4 Turbo is unable to complete the conversation following a sequential instruction within a single interaction. Based on the specific step of summarization you are in, we employ one of three designated prompts:

\textit{\textbf{Scenario 7---start prompt}}: 

\textbf{Here is one example to show GPT-4 Turbo to initiate a conversation when the first document is irrelevant:} 

[{"role": "user", "content": "Could you give a summarization regarding ChatGPT application in Finance by summarizing the documents one-by-one? There are total 5 documents to summarize"}, {"role": "assistant", "content": "In the process of summarizing documents one by one. [Retrieval]$\langle$ paragraph $\rangle$ First Document $\langle /$ paragraph $\rangle$[Irrelevant][Context] No context till now [/Context] The retrieval text is not relevant with the user's topic $\langle$ 4 documents left to summarize $\rangle$[Topic] user's topic"}] 

\textbf{Here is one example to show GPT-4 Turbo to initiate a conversation when the first document is relevant:}

[{"role": "user", "content": "Could you give a summarization regarding ChatGPT application in Finance by summarizing the documents one-by-one? There are total 5 documents to summarize"}, {"role": "assistant", "content": "In the process of summarizing documents one by one. [Retrieval]$\langle$ paragraph $\rangle$ First Document $\langle /$ paragraph $\rangle$[Relevant][Context] No context till now [/Context] Summarization of First Document $\langle$ 4 documents left to summarize $\rangle$[Topic] user's topic"}] 

\textbf{Here is the instruction to tell GPT-4 Turbo to initiate a conversation:} 

You are instructed to construct the start of the conversation between the assistant and the user requires the assistant to do summarization document by document on some topic.
    Some special tokens need to be added to the conversation. You are required to follow the format of the provided example, 
    including the position of special tokens. Here are special tokens: 
    
    [1] The information inside [Context] and [/Context] should be No context till now (You could diversify this sentence) 
    
    [2] [Relevant] and [Irrelevant] are used to check whether the retrieval text inside $\langle$paragraph$\rangle$ and $\langle /$paragraph$\rangle$ are relevant with the user query. 
    
    [3] Content inside $\langle$Count$\rangle$ and $\langle /$Count$\rangle$ is to check how many documents left to summarize. 
    
    [4] [Topic] are used to keep the topic of the user query. 
    
    The summarization should be appended after [/Context]. The retrieval text at each step should be inside of $\langle$paragraph$\rangle$ and $\langle /$paragraph$\rangle$. 
    
    Here is a relevant example: \textbf{Relevant example shown above}
    
    Here is a not relevant example: \textbf{Irrelevant example shown above}
    
    $\#\#\#$ 
    
    Now, you are instructed to follow the above examples to create the start of the convseration. There are total \textbf{5} documents, the topic is \textbf{xx}, and the first document is following: 
    
    \textbf{Content of First Document}
    
    $\#\#\#$ 
    
    The response must only be a list of four dictionaries without saying any other things.

\textit{\textbf{Scenario 7---mid prompt}}: 

\textbf{Here is one example to show GPT-4 Turbo to create the middle part of the conversation}:

[{"role": "assistant", "content": "In the process of summarizing documents one by one. [Retrieval]$\langle$ paragraph $\rangle$ First Document $\langle /$ paragraph $\rangle$[Relevant][Context] No context till now [/Context] Summarization of First Document $\langle$ 4 documents left to summarize $\rangle$[Topic] user's topic"}, {"role": "assistant", "content": "In the process of summarizing documents one by one. [Retrieval]$\langle$ paragraph $\rangle$ Second Document $\langle /$ paragraph $\rangle$[Irelevant][Context] Summarization of First Document  [/Context] Summarization of First Document $\langle$ 3 documents left to summarize $\rangle$[Topic] user's topic"}] 

\textbf{Here is the instruction to tell GPT-4 Turbo to create the middle part of the conversation} 

You are instructed to construct the conversation between the assistant itself and its goal is to do summarization document by document on some topic. Some sepcial tokens need to be added to the convseration.
    You are required to follow the format of the provided example, including the position of special tokens. Here are special tokens: 
    
    [1] The information inside [Context] and [/Context] is the context you need to rely on when you do the summarization by combining with the retrieval text. 
    
    [2] [Relevant] and [Irrelevant] are used to check whether the retrieval text inside $\langle$paragraph$\rangle$ and $\langle /$paragraph$\rangle$ are relevant with the user query. 
    
    [3] Content inside $\langle$Count$\rangle$ and $\langle /$Count$\rangle$ are to check how many documents left to summarize. 
    
    [4] [Topic] are used to keep the topic of the user query. 
Here is one example: 

    \textbf{One-shot example shown above}
    
    $\#\#$Now, I will provide you with the first piece of the conversation. You need to keep it UNCHANGED. Here is the first piece of the convseration: 
    
    \textbf{First Piece of the Conversation}$\#\#$

    and here is the new retrieval text: 

    \textbf{New Retreivel text to be processed}$\#\#$

    $\#\#$Construct the new piece of the conversation: Context should keep unchanged if [Irrelevant] appears on the first piece of conversation and need to be changed to the summarization in the first piece if the [Relevant] appears in the first piece of conversation.
    If the new retrieval text is still irrelevant to the user query, the summarization should be same as the context; if it is relevant, then the summarization should consider both the content of context and the retrieval text (DO NOT LOSE ANY INFORMATION IN THE CONTEXT)$\#\#$

    $\#\#$The position of summarization should be appended after $\langle /$Context $\rangle$
    !!!!(DO NOT LOSE ANY INFORMATION IN THE CONTEXT EVEN EXTENDING THE LENGTH OF THE SUMMARIZATION. IT IS VERY IMPORTANT)!!!! 
    You MUST RETURN ME A LIST OF TWO DICTIONARIES WITHOUT SAYING ANY OTHER THINGS$\#\#$

\textit{\textbf{Scenario 7---end prompt}}: 

\textbf{Here is one example to show GPT-4 Turbo to end a conversation:} 

[{"role": "assistant", "content": "In the process of summarizing documents one by one. [Retrieval]$\langle$ paragraph $\rangle$ Second Document $\langle /$ paragraph $\rangle$[Irelevant][Context] Summarization so far [/Context] Final Summarization $\langle$ 0 documents left to summarize $\rangle$[Topic] user's topic"}, {"role": "assistant", "content": "Here is the final summarization: Final Summarization"}] 

\textbf{Here is the instruction to tell GPT-4 Turbo to end the conversation} 

You are instructed to construct the final step of the onversation between the assistant itself and its goal is to do summarization document by document on some topic. Some sepcial tokens need to be added to the convseration.
    You are required to follow the format of the provided example, including the position of special tokens. Here are special tokens: 
    
    [1] The information after [Context] should be the information you should not forget when you do the summarization. 
    
    [2] [Relevant] and [Irrelevant] are used to check whether the retrieval text inside <paragraph> and $\langle /$paragraph$\rangle$ are relevant with the user query. 
    
    [3] Content inside $\langle$Count$\rangle$ and $\langle/$Count$\rangle$ are to check how many documents left to summarize.  
    
    [4] [Topic] are used to keep the topic of the user query.  
    
    Here is one example: 

    $\#\#\#$ \textbf{One-shot example shown above} $\#\#\#$ 

     $\#\#$Now, I provide you with the first piece of convseration. You need to keep it UNCAHNGED. Here is the first piece of the conversation: 
     
    \textbf{First piece of the conversation} 
    $\#\#$ 
    
    $\#\#$All you need to do is to generate next piece of conversation. If [Irrelvant] appears in the first piece, the final summarization is the context; if [Relevant] appears, the final summarization is the summarization from last part, which is the content after $\langle /$Context $\rangle$ $\#\#$
    
    You need to return me both the first piece and your generated conversation. You MUST RETURN ME A LIST OF TWO DICTIONARIES WITHOUT SAYING ANY OTHER THINGS.

\subsection{Evaluation Prompts for Benchmarking Other Large Language Models in Experiments}
\label{app: eval prompts}

\textit{\textbf{Zero-shot Prompts for Aspect 1}}:  

[INST]
    You are given following task based on the given text. Here is the text: $\#\#\#$\textbf{text}$\#\#\#$. You first check if the text is relevant with the 
    $\#\#\#$\textbf{user\_topic}$\#\#\#$. Using following answer format: 
    
    (1) if it is relevant, return me using following format "The retrieval text is relevant with user's topic. Start summarizing on it: content of the summarization"  
    
    (2) if it is not relevant, return me using following format "The retrieval text is not relevant with user's topic" 
    
    Here is your reply: [/INST] 

\textit{\textbf{Chain of Thought Prompts for Aspect 1}}:  

[INST]
    You are given following task based on the given text. Here is the text: $\#\#\#$\textbf{text}$\#\#\#$. You first check if the text is relevant with the 
    $\#\#\#$\textbf{user\_topic}$\#\#\#$.  You are required to finish the task step by step: 
    
    The first step is to determine the relevancy of the retrieval text to the user topic. 
    
    Then the second step is based on the result of the relevancy:  
    
    (1) if it is relveant, return me using following format "The retrieval text is relevant with user's topic. Start summarizing on it: content of the summarization" 
    
    (2) if it is not relevant, return me using following format "The retrieval text is not relevant with user's topic" 
    
    Here is your reply: [/INST] 

\textit{\textbf{One-shot Chain of Thought Prompts for Aspect 1}}: 

[INST]
    You are given following task based on the given text. Here is the text: $\#\#\#$\textbf{text}$\#\#\#$. You first check if the text is relevant with the 
    $\#\#\#$\textbf{user\_topic}$\#\#\#$.  You are required to finish the task step by step: 
    
    The first step is to determine the relevancy of the retrieval text to the user topic. 
    
    Then the second step is based on the result of the relevancy:  
    
    (1) if it is relveant, return me using following format "The retrieval text is relevant with user's topic. Start summarizing on it: content of the summarization" 
    
    (2) if it is not relevant, return me using following format "The retrieval text is not relevant with user's topic" 

    Here is an example: 
    
    The user topic would like to know the summary about ChatGPT application in Finance. The retrieval text is $\#\#\#$ ChatGPT is a chatbot developed by OpenAI and launched on November 30, 2022. Based on a large language model, it enables users to refine and steer a conversation towards a desired length, format, style, level of detail, and language $\#\#\#$. Then, you need to output it is not relevant
    since the user asks the specific finance application of ChatGPT but the retrieval text reflects the ChatGPT introduction. 
    
    Here is your reply: [/INST] 

\textit{\textbf{Zero-shot prompts for Aspect 2}}: 

[INST]
    You are a summarization assistant to summarize following text and return ONLY the summary to me.
    Here is the text $\#\#\#${user\_text}$\#\#\#$ [/INST]

\textit{\textbf{Zero-shot Prompts for Aspect 3}}:

[INST]
    You are given two tasks based on the given two texts. Here is the user text: $\#\#\#$\textbf{use\_text}$\#\#\#$. Here is the retrieval text: 
    $\#\#\#$\textbf{retrieval\_text}$\#\#\#$.You first check if the retrieval text is relevant with the user text, and if it is relevant, 
    check if there is any information conflict between the retrieval text and the user text. Using following format: 
    
    (1) if they are not relevant, you should return to me: the user text is not relevant with the retrieval text. Start summarizing only on user text: content of the summarization 
    
    (2) if they are relevant but the retrieval text has information conflict with the user text, you only need to return "There is information conflict between the user text and the retrieval text"  
    
    (3) if they are relevant and there is no information conflict between them, you should return to me: the user text is relevant with the retrieval text and there is no information conflict. Start summarizing on 
    both retrieval text and user text: content of the summarization. 
    [/INST]

\textit{\textbf{Chain of Thought Prompts for Aspect 3}}:

[INST] 
    You are instructed to finish following text step by step. Here is the user text: $\#\#\#${user\_text}$\#\#\#$.  
    Here is the retrieval text: 
    $\#\#\#${retrieval\_text}$\#\#\#$. 
    The first step is to check if the retrieval text is relevant with the user text. Based on the check result, you are ready to implement the following step. 
    (1) if they are not relevant, you should return to me: the user text is not relevant with the retrieval text. Start summarizing only on user text: content of the summarization 
    
    (2) if they are relevant but the retrieval text has information conflict with the user text, you only need to return "There is information conflict between the user text and the retrieval text" 
    
    (3) if they are relevant and there is no information conflict between them, you should return to me: the user text is relevant with the retrieval text and there is no information conflict. Start summarizing on 
    both retrieval text and user text: content of the summarization. 
    [/INST] 

\textit{\textbf{One-shot Chain of Thought Prompts for Aspect 3}}: 

[INST] 
    You are instructed to finish following text step by step. Here is the user text: $\#\#\#${user\_text}$\#\#\#$.  
    Here is the retrieval text: 
    $\#\#\#${retrieval\_text}$\#\#\#$. 
    The first step is to check if the retrieval text is relevant with the user text. Based on the check result, you are ready to implement the following step. 
    (1) if they are not relevant, you should return to me: the user text is not relevant with the retrieval text. Start summarizing only on user text: content of the summarization 
    
    (2) if they are relevant but the retrieval text has information conflict with the user text, you only need to return "There is information conflict between the user text and the retrieval text" 
    
    (3) if they are relevant and there is no information conflict between them, you should return to me: the user text is relevant with the retrieval text and there is no information conflict. Start summarizing on 
    both retrieval text and user text: content of the summarization. 

    Here is one example: The user text is $\#\#\#$ The Ragdoll is a breed of cat with a distinct colorpoint coat and blue eyes. Its morphology is large and weighty, and it has a semi-long and silky soft coat. American breeder Ann Baker developed Ragdolls in the 1960s. They are best known for their docile, placid temperament and affectionate nature.
    $\#\#\#$ 
    
    The retrieval text is$\#\#\#$A domestic short-haired cat is a cat possessing a coat of short fur, not belonging to any particular recognised cat breed. In the United Kingdom, they are colloquially called moggies.$\#\#\#$ Then, in this example,
    your reply should be: The user text is not relevant with the retrieval text. Start summarizing only on user text: Ragdolls are large, gentle cats with colorpoint coats and blue eyes.[/INST]

\subsection{Evaluation Prompts for Benchmarking Other Summarization Frameworks} 
\label{app: multi} 

--------------------------------------- 

\textit{\textbf{Prompts for Stuff summarization}} 

Write a summary of the following text regarding topic \textbf{topic} and skip irrelevant text with respect to this topic. 

            Here is the text: \textbf{text} 
            
--------------------------------------- 

\textit{\textbf{Prompts for Map-Reduce summarization}}

\textbf{Map prompt}:

Write a summary of this chunk of text regarding topic \textbf{topic} that includes the main points and any important details (skip irrelevant text with respect to this topic.) 

Here is the text: \textbf{text} 

\textbf{Reduce prompt}:

Write a concise summary of the following text delimited by triplet backquotes. ```\textbf{text}``` 

                Here is your summary: 

--------------------------------------- 

\textit{\textbf{Prompts for Refine summarization}}  

\textbf{Question prompt}:

Provide a summary of the following text with respect to topic \textbf{topic} (skip irrelevant text with respect to the topic): 

            TEXT: \textbf{text} 
            
            SUMMARY: 

\textbf{Refine prompt}:

Write a concise summary of the following text delimited by triple backquotes. 

                    ```\textbf{text}``` 
                    
                    SUMMARY:

\newpage
\subsection{Transformation Text for Special Tokens and System Prefixes} 
\label{app:transform_text}

\begin{table}[!h]
\centering
\footnotesize
%
%
\begin{tblr}{
  colspec={Q[l,b] X[l,b]},
  row{2,4,6,8,10,13,15}={bg=azure9},
  row{3,5,7,9,11,14,16}={bg=gray9},
  row{1,12}={c,bg=gray4,fg=white},
}
Special Tokens & \SetCell[c=1]{c} Natural Text Alternatives\\
\textbf{[Retrieval] $|$ [No Retrieval]} & Here is the retrieval text to be used for summarization $|$ There is no need to retrieve since user provides its own text \\
\textbf{[Relevant]$|$ [Irrelevant]} & The retrieval text is relevant $|$ irrelevant with the user's text \\
\textbf{[Continue to Use User's Text]} & The retrieval text is not relevant with the user's text. Ignore it and use the user's text to form the summarization as follows: \\
\textbf{[Information Conflict]} & Although the retrieval text is relevant with user's text, there is an information conflict between user's text and the retrieved text. \\
\textbf{[Augmenting User's Text]} & The retrieval text is relevant with user's text. MUST COMBINE user's text and retrieved text to write the final summarization. \\
\textbf{[Context]} & Context to be used for the summarization\\ 
\textbf{[/Context]} &  If the retrieval text is not relevant with the user's topic, keep the summarization at this step same as the context;\\ 
\textbf{[/Context]} & If the retrieval text is relevant with the user's topic, combine retrieval text with context information. Here is the summarization at this step:  \\
\textbf{$<$Count$>$, $<$/Count$>$}& Start counting how many documents left to summarize. Current summarization step you are at: \\
\textbf{[Topic]} & Here is the topic to be kept to check if retrieval text is relevant with the user's query:  \\
Aspect &\SetCell[c=1]{c} System Prefix \\
\textbf{1} & You are a summarization assistant to retrieve the text based on user's topic and then do the summarization. \\
\textbf{2} & You are a summarization assistant to do the summarization based on user's text. \\
\textbf{3} & You are a summarization assistant to decide if combining the retrieval text with user's text to do the summarization based on its relevancy: \\
\textbf{4} & You are a summarization assistant to summarize the documents one by one. \\
\end{tblr}
\vspace*{-1em}
\end{table}

\end{document}